\documentclass{article}
\usepackage{graphicx} 
\usepackage{wrapfig}
\usepackage{authblk} 
\author[1]{Hangyi Shen}
\author[1]{Yizhi Pan}
\author[1]{Tiansuo Li}
\author[1]{Weiqi Jiang}
\author[1]{Guanqun Sun}
\affil{\textsuperscript{1}Hangzhou Medical College, Hangzhou, China}
\usepackage{placeins}
\usepackage{microtype}
\usepackage{array}
\usepackage{float}
\usepackage{subcaption}
\graphicspath{{./photo/}}
\usepackage{amsmath}
\usepackage{booktabs} % 专用于绘制高质量的三线表
\usepackage{multirow} % 用于处理跨行合并
\usepackage{caption}
\title{Image Feature Fusion-based Federated Client Unlearning (FCU) }

\date{May 2026}
\begin{document}
\maketitle
\begin{abstract}
Major data protection regulations all mention the “right to be forgotten,” and that’s what pushed federated unlearning (FU) techniques forward. But one stubborn issue remains: catastrophic forgetting—you erase the target knowledge, yet somehow you also end up throwing out essential retained knowledge, which then hurts the model’s global generalization.

To get a better balance between unlearning effectiveness and generalization ability, we propose something called Image Feature Fusion‑based Federated Client Unlearning (IFF‑FCU). The idea is to bring in a linear Image Feature Fusion mechanism (Mixup) that dynamically creates mixed samples, bridging the gap between forget‑distribution and retain‑distribution. What this strategy does isn’t just deleting a few discrete data points—it theoretically widens and regularizes the forgetting boundary.

We ran extensive experiments on medical imaging benchmarks (RSNA‑ICH and ISIC2018), and the results show that our approach achieves reasonably good unlearning. For instance, on the ICH dataset, IFF-FCU achieves a highly competitive Error deviation from the retrained gold standard, demonstrating robust improvements over existing baselines.

\vspace{1em}
\noindent \textbf{keywords:}Medical Imaging;Catastrophic Forgetting;Mixup;Image Feature Fusion;Frequency-Guided Memory Preservation;Model-Contrastive Unlearning;Federated Unlearning
\end{abstract}
\section{Introduction}

Federated Learning (FL) has emerged as a transformative privacy-preserving paradigm in medical imaging by enabling collaborative model training among multiple parties without sharing raw patient data. However, the trained global model may still implicitly memorize the contributions of specific clients, conflicting with the "right to be forgotten" mandated by regulations such as GDPR\cite{voigt2017eu} and CCPA\cite{harding2019understanding}. Consequently, how to efficiently eliminate the influence of a target client without costly retraining from scratch has become a research hotspot in the field of Federated Unlearning (FU). Looking across current research, despite progress in federated unlearning, it still commonly suffers from the dilemma of residual feature entanglement and catastrophic forgetting. Existing paradigms mostly confine the unlearning operation to isolated forgetting samples. Building upon recent advances in federated unlearning, this paper explores the transition from optimizing isolated data points to regularizing continuous feature boundaries. This analysis from the perspective of feature space entanglement lays the theoretical foundation for the image feature fusion mechanism proposed in IFF-FCU.

\section{Related Work}
Early studies attempted to achieve "exact unlearning" by manipulating historical gradients to avoid the high cost of retraining from scratch.FedEraser\cite{liu2021federaser} was a pioneer along this path, which iteratively removes the contribution of a target client by exploiting historical parameter updates stored on the server side.But,such methods require additional storage space on the server and carry the risk of data reconstruction by a malicious server using historical data.To break free from reliance on historical parameters, research shifted toward "approximate unlearning," i.e., directly applying reverse updates or penalty terms to the current model.FFMU\cite{che2023fast} combines stochastic gradient smoothing and quantization to perform the unlearning operation. UPGA\cite{halimi2022federated} formulates unlearning as a constrained gradient ascent problem aimed at maximizing the empirical loss of the target client. To restore performance, FUKD\cite{wu2022federated} introduces performance repair based on knowledge distillation; whereas MoDe\cite{zhao2023federated} utilizes pseudo-labels generated by a "Bad Teacher" model to guide knowledge erasure.Such methods frequently face a "utility–thoroughness" trade-off. FFMU\cite{che2023fast} tends to exhibit insufficient unlearning, while MoDe\cite{zhao2023federated} and UPGA\cite{halimi2022federated}, although capable of producing high unlearning error, often lead to severe degradation in global performance, i.e., "over-forgetting."
Addressing the limitation that output-layer distillation may not fully capture deep feature representations, the FCU\cite{deng2024enable} framework introduces feature-level unlearning interventions. The FCU\cite{deng2024enable} framework proposed by Deng et al.\ introduced Model Contrastive Unlearning (MCU), which forces the model to output features different from the trained global model while performing similarly to a degraded model that has never encountered the forgetting data. Meanwhile, a Frequency-Guided Memory Protection (FGMP) mechanism utilizes FFT to transform parameters into the frequency domain, retaining low-frequency global knowledge and erasing high-frequency local knowledge.While FCU\cite{deng2024enable} runs 10–15 times faster than full retraining, its MCU operation still only handles discrete target forgetting samples, which means some cleanup blind spots could remain along complex feature boundaries.

Looking at recent work, FedzMuGAN\cite{maher2026fedzmugan} tackles the “client offline” issue in federated unlearning by having the server use a Generative Adversarial Network (GAN) to create pseudo data that mimics the categories to be forgotten—this way model reconstruction can finish without needing the client. But the catch is that unlearning thoroughness depends entirely on how good those synthetic data are, and GAN training stability itself remains tricky.

Jellyfish\cite{wang2026jellyfishzeroshotfederatedunlearning} brings in a Knowledge Disentanglement mechanism: it splits client‑specific contributions from global common knowledge in the feature space, which enables efficient zero‑shot unlearning. That approach does great things for preserving model accuracy, yet the disentanglement algorithm comes with high computational costs.

Then there’s FUSED\cite{zhong2025unlearning} proposing reversible unlearning based on Selective Sparse Adapters—by checking each layer’s sensitivity to knowledge, it builds lightweight adapters only in critical layers. RobustFU\cite{sheng2024robust}, in contrast, focuses on defense: a robust aggregation mechanism ensures security in case malicious clients try to exploit the unlearning operation to inject backdoors, though that does eat into communication efficiency to some degree.

After comparing existing literature, we notice that most current algorithms assume the forgetting domain and the retention domain can be separated cleanly in the feature space. However, when you look at high‑dimensional medical imaging manifolds, that assumption simply doesn’t hold. The knowledge of the forgetting data and that of the retained data are highly entangled in the feature space.

This finding explains the reason for the catastrophic forgetting observed in MoDe\cite{zhao2023federated} and UPGA\cite{halimi2022federated}: merely repelling isolated forgetting points can tear apart adjacent retention features. Conversely, if the intervention is too weak, it cannot cover the ``gray zone'' at the feature boundary. Therefore, efficient unlearning must shift from isolated point optimization to the regularization of the entire feature manifold boundary.

\section{Materials and Methods}
\subsection{System Architecture}
Following the standard protocol established in existing literature \cite{deng2024enable}, our system comprises a central aggregator and five distributed nodes. We denote the set of these participating clients as $C = \{C_1, C_2, C_3, C_4, C_5\}$, with their respective localized datasets defined sequentially as $D = \{D_1, D_2, D_3, D_4, D_5\}$.

\begin{figure}[htbp]
  \centering
  \includegraphics[width=0.85\textwidth]{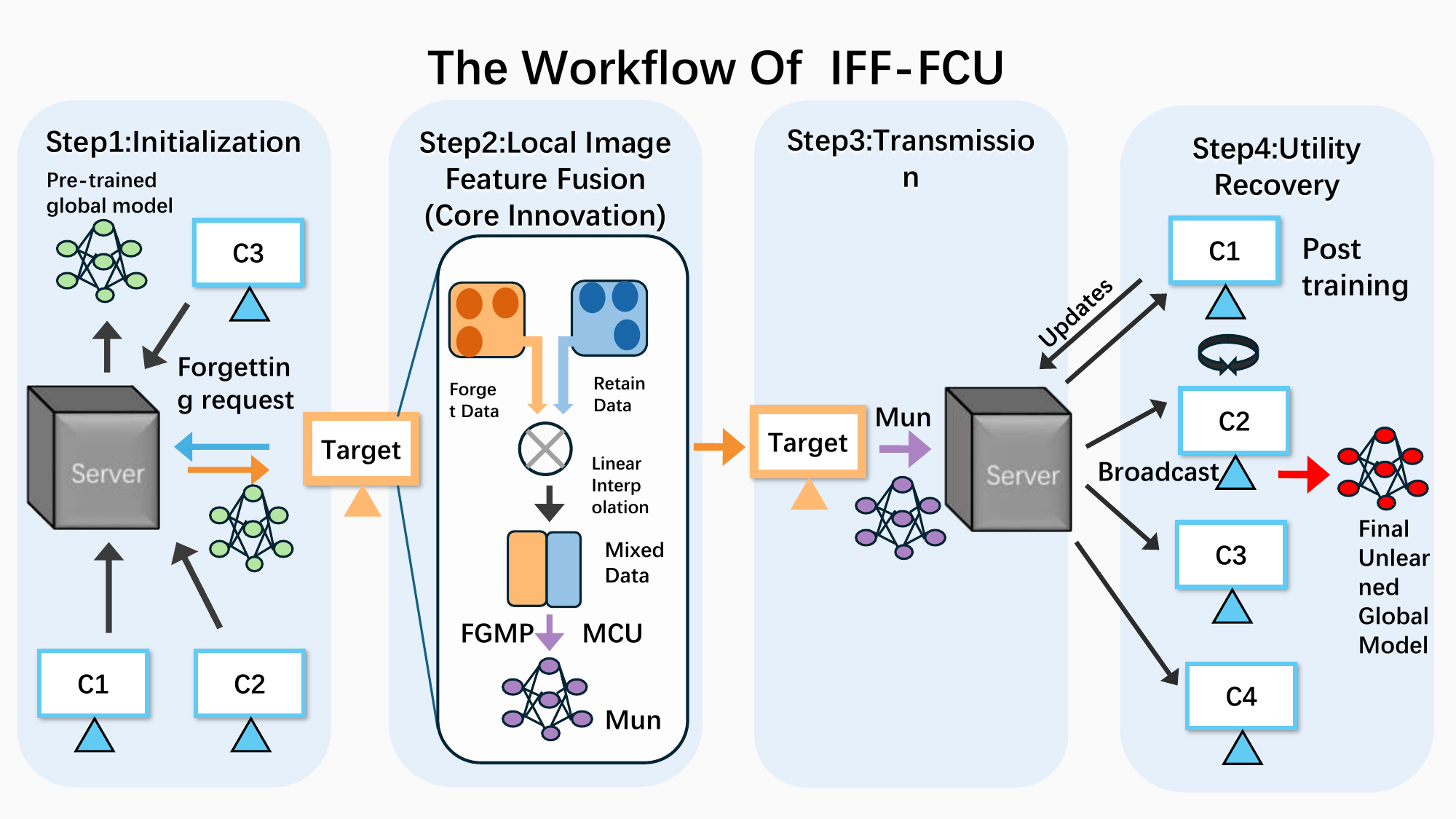}
  \caption{Overview of the proposed IFF-FCU framework. At the local level, the designated client executes the unlearning procedure utilizing a Mixup strategy (performing linear interpolation across the retained and forgotten data). Subsequently, the sanitized model parameters are uploaded to the central server, establishing the baseline weights for the ensuing post-training phase across all other participating nodes.}
  \label{fig:IFF-FCU}
\end{figure}

The overall workflow is as follows. After \( t \) rounds of federated learning, each client possesses a trained global model \( M_{\text{tr}} \). A client \( C_i \) that wishes to withdraw, referred to as the target client, requests the removal of the contribution of its data \( D_u \) from \( M_{\text{tr}} \). The goal of this study is to perform unlearning on the Mixup samples — the linear interpolation of retain samples and forget samples — thereby effectively eliminating the influence of the target forget dataset on \( M_{\text{tr}} \) and producing the unlearned model \( M_{\text{un}} \). The proposed IFF-FCU framework is illustrated in Figure~\ref{fig:IFF-FCU}. Before performing local unlearning on the target forget dataset, the target client first conducts a one-to-one linear interpolation fusion between each sample to be forgotten from its own dataset and a sample from the retain dataset obtained from other participating clients. The resulting mixed samples are then fed into the model, which performs unlearning through a combination of contrastive learning (MCU) and frequency-guided memory preservation (FGMP).
\subsection{Image Image Feature Fusion Mechanism}
The intuition of our Image Feature Fusion Mechanism is to expand the scope of unlearning by creating synthetic samples that bridge the gap between the forget data and retain data, thereby encouraging the unlearned model to generalize its forgetting behavior beyond the isolated forget samples. While MCU directly applies contrastive learning on raw forget samples—which may lead to overfitting on limited forget data and insufficient forgetting coverage—we propose to fuse features from both forget and retain samples at the input level, allowing the model to learn a smoother and more generalized unlearning boundary.

Specifically, given a forget sample \(x_f\) from the unlearning dataset \(\mathcal{D}_u\) and a retain sample \(x_r\) from the remaining data, we create a mixed sample through linear interpolation:
\[
x_{\text{mix}} = \lambda \cdot x_f + (1-\lambda) \cdot x_r
\]
where the mixing coefficient \(\lambda\) is sampled from a Beta distribution \(\lambda \sim \text{Beta}(\alpha, \alpha)\). The mixed sample \(x_{\text{mix}}\) inherits characteristics from both the forget and retain distributions, effectively creating a continuum of samples that gradually transition from the forget domain to the retain domain

The key insight behind this mechanism is two-fold: (1) Expanded forgetting scope—by unlearning on mixed samples that incorporate retain features, we prevent the model from overfitting to specific forget samples and instead encourage it to forget a broader region of the feature space surrounding the forget data; (2) Preservation of utility—the retain component in the fusion acts as an anchor, ensuring that the unlearning process does not catastrophically interfere with the model's performance on the retain set.

Our Image Feature Fusion Mechanism is integrated with the Model-Contrastive Unlearning framework, where the contrastive loss is computed on the mixed features:
\[
\mathcal{L}_{\text{fusion}} = -\log\frac{\exp(\text{sim}(z_{\text{mix}}, z_{\text{down}})/\tau)}{\exp(\text{sim}(z_{\text{mix}}, z_{\text{down}})/\tau) + \exp(\text{sim}(z_{\text{mix}}, z_{\text{tr}})/\tau)}
\]
where \(z_{\text{mix}}\) denotes the feature extracted from the mixed sample \(x_{\text{mix}}\). By pulling \(z_{\text{mix}}\) toward the downgraded model \(M_{\text{down}}\) and pushing it away from the trained model \(M_{\text{tr}}\), we effectively transfer the forgetting effect from the mixed samples back to the forget distribution while maintaining model utility through the retain component. $L_{total} = L_{fusion}$.

\paragraph{Note: Achieving Regularization through Retention Anchoring and Boundary Broadening.} 
Recent progress in representation learning gives solid theoretical backing to using feature fusion at the input level for widening the unlearning boundary. In high‑dimensional medical imaging, what really drives catastrophic forgetting is the “catastrophic overlap” between the retain dataset’s feature manifold and the forget dataset’s feature manifold. Peng\cite{peng2025adversarial}et al. (2025) show in their recent theoretical work that synthesizing mixed samples through interpolation can actually simulate those heavily entangled overlapping regions. Instead of optimizing over isolated discrete points, our Mixup mechanism optimizes the unlearning objective over this synthesized continuous space, which effectively expands the forgetting boundary.

One thing worth noting: optimizing $\mathcal{L}_{fusion}$does not depend on the original task ‐-specific class labels. So we explicitly steer the unlearning direction by generating dynamic pseudo‑labels driven by the fusion ratio. In concrete terms, the interpolation coefficient $\lambda$ acts as a decisive threshold for the alignment of the representation:

\begin{itemize}
    \item \textbf{Retention Anchoring ($\lambda \leq 0.5$):} If the mixed sample $x_{mix}$ is mainly made up of the retaining feature $x_{r}$,we assign it a contrastive pseudo‑label of 0. This tells the objective function to pull the mixed representation hard towards the output of the global model trained before ‐ ($h_{s_0}$). In this case, the pre‑trained global model works as a clear retaining anchor—it forces the current model to copy its past predictions on the retained class distribution, which physically protects global utility.
    
    \item \textbf{Robust Erasure ($\lambda > 0.5$):}Now flip the situation: when the forgetting feature $x_{f}$takes over, the pseudo‑label becomes 1.That actively pulls the representation toward the downgraded model($h_{s_1}$)to get a robust erasure.
\end{itemize}

Using this $\lambda$-controlled deterministic switching mechanism, IFF‑FCU draws a line in the latent space between the forgetting manifold and the retaining manifold, and it does all that without needing any ground‑truth semantic labels.

\subsection{Overall Workflow of the IFF-FCU Framework}

Within the IFF‑FCU framework, the whole process—starting from initial model training all the way to forgetting sensitive data—breaks down into four key stages. One core piece of our design is that we bring in the Mixup‑based Image Feature Fusion mechanism during the local unlearning stage.

\begin{enumerate}
    \item \textbf{Federated Pre-training and Unlearning Initialization.} \\
    The server starts by getting all clients to run standard federated pre‑training together, which produces an initial model that carries globally generalizable knowledge. Then, once the system receives a data deletion request from a specific client (called the “forget client”), the server triggers the unlearning procedure and sends the current pre‑trained model over to that client to act as the unlearning baseline.

    \item \textbf{Mixup-Enhanced Local Contrastive Unlearning (Core Innovation).} \\
    This is the central component of our framework. Unlike traditional contrastive unlearning methods that exclusively target the forgotten data, we introduce the Mixup Image Feature Fusion mechanism to reshape the model's feature space. The specific process is as follows:
    \begin{itemize}
        \item \textit{Constructing the Mixed Transition Zone:} In each training iteration, the target unlearning client needs to extract not only the forget samples, but also request retain samples from other clients via the server. Through the Mixup mechanism, the system dynamically interpolates the two, generating ``mixed samples'' that contain both forgotten and retained features. 
        \item \textit{Expanding the Forgetting Boundary:} Subsequently, the contrastive learning module operates on these mixed samples. It drives the model to push away from the pre-trained state (the forgetting target) on the mixed data, while pulling close to the downgraded model.
    \end{itemize}

    \item \textbf{Frequency-Guided Local Knowledge Preservation.} \\
    Concurrently with the Mixup contrastive unlearning, the system periodically applies Frequency-Guided Memory Preservation (FGMP). It filters and preserves the low-frequency global knowledge of the pre-trained model in the frequency domain, only allowing the model to forget the high-frequency local information associated with the specific client.

    \item \textbf{Global Model Post-training and Recovery.} \\
    Once local unlearning is complete, a ``clean model'' devoid of the specific client's memory is transmitted back to the server. The server then broadcasts it to all remaining healthy clients. These clients utilize their respective retained data to perform a small number of Federated Averaging (FedAvg)\cite{mcmahan2017communication} fine-tuning rounds on this clean model.
\end{enumerate}
\FloatBarrier
\section{Results}
\subsection{Experimental implementation}

\noindent \textbf{Datasets.}Our empirical evaluations are conducted on two standard medical imaging benchmarks: the RSNA-ICH\cite{flanders2020rsna} and the ISIC2018(HAM10000) skin lesion corpus \cite{tschandl2018ham10000}. In alignment with the experimental protocol established by \cite{deng2024enable}, we partition the data into training, validation, and testing subsets adopting a 70\%, 10\%, and 20\% split, respectively. To simulate a heterogeneous federated environment, the training subset is allocated across five discrete local nodes guided by a Dirichlet distribution ($\alpha = 1.0$).

\noindent \textbf{Implementation Details.}For both classification tasks, we use DenseNet121 as the main feature extractor. Model weights get updated via the Adam optimizer, with exponential decay rates set to $\beta_1 = 0.9$ and $\beta_2 = 0.99$. We also apply a differential learning rate scheme: the specific client that performs unlearning runs at $1 \times 10^{-5}$, while all other federated participants use a step size of $1 \times 10^{-4}$. A fixed batch size of 64 is used across every local computation. The local erasure process runs for exactly 100 iterations. The MCU loss uses a temperature of $\tau = 0.5$. The frequency-guided memory preservation (FGMP) module gets triggered every $T_{\text{FGMP}} = 10$ steps. For the interpolation mechanism, we set the execution probability $p_{\text{mixup}}$ to 0.5 and the shape parameter $\alpha_{\text{mixup}}$ to 0.2. During the global utility recovery phase, the system goes through 10 rounds of server-client communication, each round including 20 local fine-tuning steps. All input images are resized to $224 \times 224$ pixels. To improve generalization, the first task uses a full set of spatial transforms (random flips, rotations, translations, scaling) plus Gaussian blurring. The second task, on the other hand, sticks to flips, rotations, and translations only, following exactly the preprocessing pipeline described in \cite{deng2024enable}.

\noindent \textbf{Software and Hardware Environment:} All experiments run under Python 3.9.23 with PyTorch 1.8.0 + cu111. Training and unlearning are accelerated by an NVIDIA GeForce RTX 2080 Ti (22GB VRAM) to handle the heavy compute from high-resolution medical images.

\noindent \textbf{Evaluation Metrics.} We evaluate our framework from two main perspectives:
\begin{itemize}
    \item \textbf{Utility (Fidelity):} This looks at how well the model keeps its performance on the data that stays behind. The metrics we use include F1 score, Accuracy, and the misclassification rate $\text{Error}_r$---computed over the preservation subset $\mathcal{D}_r$ (that's the data belonging to non-target nodes $\mathcal{C} \setminus \{C_i\}$), plus the standard testing error $\text{Error}_t$.
    
    \item \textbf{Unlearning Efficacy:} This measures how completely the departing node's data gets erased. We quantify this using the specific misclassification penalty $\text{Error}_f$, which is evaluated only on the target client's own local data $\mathcal{D}_f$.
    
    \item \textbf{Efficiency:} We measure efficiency by how long it takes for the model to get back to its best performance---in other words, the convergence time.
\end{itemize}
\subsection{Comparison with Baselines}
To see how stable and effective the proposed Image Feature Fusion paradigm really is, we ran a fairly broad set of comparisons against several reference points. First, we set the performance boundaries by comparing against three things: the original global model (before any unlearning), a simple finetuning baseline, and a model retrained completely from scratch—that last one serves as the theoretical ideal for perfect memory removal. Second, to check whether our architectural design actually helps, we compare against the underlying framework we directly build on, which is FCU \cite{deng2024enable}. Third, to get a sense of where we stand in the wider machine unlearning field, we pit our method against five current top‑performing federated unlearning techniques: FFMU \cite{che2023fast}, MoDe \cite{zhao2023federated}, UPGA \cite{halimi2022federated}, FedEraser \cite{liu2021federaser}, and FUKD \cite{wu2022federated}.
\begin{table}[H]
\centering
\caption{ICH*the retrained model serves as the gold standard.}
\label{tab:1}
\begin{tabular}{c|cccc|c|c}
\toprule
\multirow{2}{*}{Methods} & \multicolumn{4}{c|}{Fidelity} & \multicolumn{1}{c|}{Efficacy} & \multicolumn{1}{c}{Efficiency} \\
\cmidrule{2-7}
& Accuracy & F1 & $Error^t$ & $Error^r$ & $Error^f$ & Runtime (s) \\
\midrule
Origin & 86.84 & 86.84 & 13.15 & 8.44 & 6.59${(-5.7)}$  & 3612 \\
Retrain & 85.52 & 85.40 & 14.48 & 10.60 & 12.28*${(0.0)}$  & 2873 \\
\midrule
Finetune & 86.11 & 86.01 & 13.88 & \textbf{7.86} & 7.25${(-5.0)}$  & 301 \\
FFMU \cite{che2023fast} & 85.09 & 84.98 & 14.90 & 10.28 & 9.79${(-2.5)}$ & 249 \\
MoDe \cite{zhao2023federated} & 80.88 & 80.62 & 19.12 & 14.59 & 18.34${(+6.1)}$  & 509 \\
UPGA \cite{halimi2022federated} & 78.96 & 78.93 & 21.04 & 20.43 & 20.25${(+8.0)}$ & 1031 \\
FedEraser \cite{liu2021federaser} & 84.32 & 84.30 & 15.67 & 11.72 & 15.29${(+3.0)}$ & 906 \\
FUKD \cite{wu2022federated} & 83.66 & 83.54 & 16.34 & 12.49 & 15.63${(+3.4)}$ &  1347 \\
\midrule
FCU\cite{deng2024enable} & \textbf{86.40} & \textbf{86.32} & \textbf{13.60} & 8.11 & 11.37${(-0.9)}$  & 177 \\
ours & 85.26 & 85.21 & 14.7  & 8.52 & \textbf{12.4}($+$0.12) & 245 \\
\bottomrule
\end{tabular}
\end{table}
\begin{table}[H]
\centering
\caption{ISIC2018}
\label{tab:2}
\begin{tabular}{c|cccc|c|c}
\toprule
\multirow{2}{*}{Methods} & \multicolumn{4}{c|}{Fidelity} & \multicolumn{1}{c|}{Efficacy} & \multicolumn{1}{c}{Efficiency} \\
\cmidrule{2-7}
& Accuracy & F1 & $Error^t$ & $Error^r$ & $Error^f$ & Runtime (s) \\

\midrule
Origin & 79.98 & 57.90 & 20.01 & 9.66 & 29.95${(-5.4)}$ & 2469 \\
Retrain & 81.52 & 54.76 & 18.47 & 5.40 & 35.37*${(0.0)}$ & 2038 \\
\midrule
Finetune & 80.52 & 55.43 & 19.47 & 8.43 & 30.49${(-4.9)}$ & 289 \\
FFMU \cite{che2023fast} & 79.67 & 44.64 & 20.33 & 10.32 & 30.99${(-4.4)}$ & 176 \\
MoDe \cite{zhao2023federated} & 73.04 & 33.63 & 26.96 & 16.55 & 50.84${(+15.5)}$  & 431 \\
UPGA \cite{halimi2022federated} & 75.08 & 38.43 & 24.91 & 17.51 & 48.28${(+12.9)}$ & 899 \\
FedEraser \cite{liu2021federaser} & 80.13 & 53.64 & 19.87 & 8.72 & 37.49${(+2.1)}$ & 672 \\
FUKD \cite{wu2022federated} & 78.11 & 41.27 & 21.89 & 12.69 & 39.72${(+4.4)}$ & 1132 \\
\midrule
FCU\cite{deng2024enable} & \textbf{81.67} & \textbf{56.32} & \textbf{18.33} & \textbf{8.22} & 34.45${(\mathbf{-0.9)}}$ & 156 \\
ours & 79.03 & 51.30 & 20.96 & 11.62 & \textbf{38.23}($+$2.86) & 120 \\
\bottomrule
\end{tabular}
\end{table}

\subsection{Results Analysis}

\noindent \textbf{Precise Unlearning Efficacy vs. Over-forgetting.} In machine unlearning, the ultimate goal is not to blindly maximize the forgotten error ($Error^f$), but rather to precisely match the distribution of a model trained from scratch on the retained dataset (the ``Retrain'' gold standard). Take baselines like MoDe\cite{zhao2023federated} and UPGA\cite{halimi2022federated}. They push $Error^f$ way too high---on ISIC2018\cite{tschandl2018ham10000}, for instance, their deviations hit $+15.5\%$ and $+12.9\%$. But that comes at a steep price: the retained data take a catastrophic performance hit, which points to serious ``over-forgetting.'' On the other side of the spectrum, methods such as Finetune and FFMU\cite{che2023fast} lean the opposite way---they tend to ``under-forget,'' showing negative deviation.

It looks like our method avoids both extremes, at least to some degree, and gives reasonably precise unlearning. On the ICH\cite{flanders2020rsna} dataset, we get an $Error^f$ of 12.4\%, which is only $+0.12$\% off the gold standard---pretty small. Over on ISIC2018\cite{tschandl2018ham10000}, the $Error^f$ comes out to 38.23\% (a $+2.86$\% deviation). That seems to indicate a fairly effective removal of the target data, and importantly, we don't see the kind of catastrophic global model collapse that happens with MoDe\cite{zhao2023federated} or UPGA\cite{halimi2022federated}. So maybe this suggests that the Image Feature Fusion Mechanism has at least some ability to gently adjust decision boundaries.

\noindent \textbf{Highly Competitive Efficiency.} Efficiency really holds back practical use of unlearning algorithms. Retraining from the ground up eats up huge compute—2873 seconds on ICH\cite{flanders2020rsna} and 2038 seconds on ISIC2018\cite{tschandl2018ham10000}, to be exact. Our method cuts that down to just 245s and 120s. On top of that, if you look at heavyweight frameworks like UPGA\cite{halimi2022federated} (1031s / 899s) and FUKD\cite{wu2022federated} (1347s / 1132s), ours runs in a fraction of their time. Sure, compared to plain FCU\cite{deng2024enable}, we do add a tiny bit of computational overhead. But given the modest progress we've made in precise unlearning alignment, that trade‑off in cost‑effectiveness doesn't seem unreasonable. One interesting thing: on ISIC2018\cite{tschandl2018ham10000}, IFF‑FCU actually ran a bit faster (120s) than FCU\cite{deng2024enable}(156s). Why? Likely because Mixup sampling, plus stochastic early stopping in some training rounds, helped each epoch converge a little faster on this dataset—its lower inter‑class variance probably helped. Still, the big takeaway here is that our method's efficiency edge over full retraining is what relatively stands out.

\noindent \textbf{Fidelity.} Achieving deep unlearning inevitably involves a natural trade-off with model fidelity. Our method maintains Accuracy and F1 scores that are highly competitive and stable, closely tracking the Retrain baseline, whereas methods aiming for extreme $Error^f$ (e.g., MoDe\cite{zhao2023federated}) suffer unacceptable fidelity losses. 
\begin{figure}[htbp]
    \centering

    \begin{subfigure}{\textwidth}
        \centering
        
        \includegraphics[width=\textwidth]{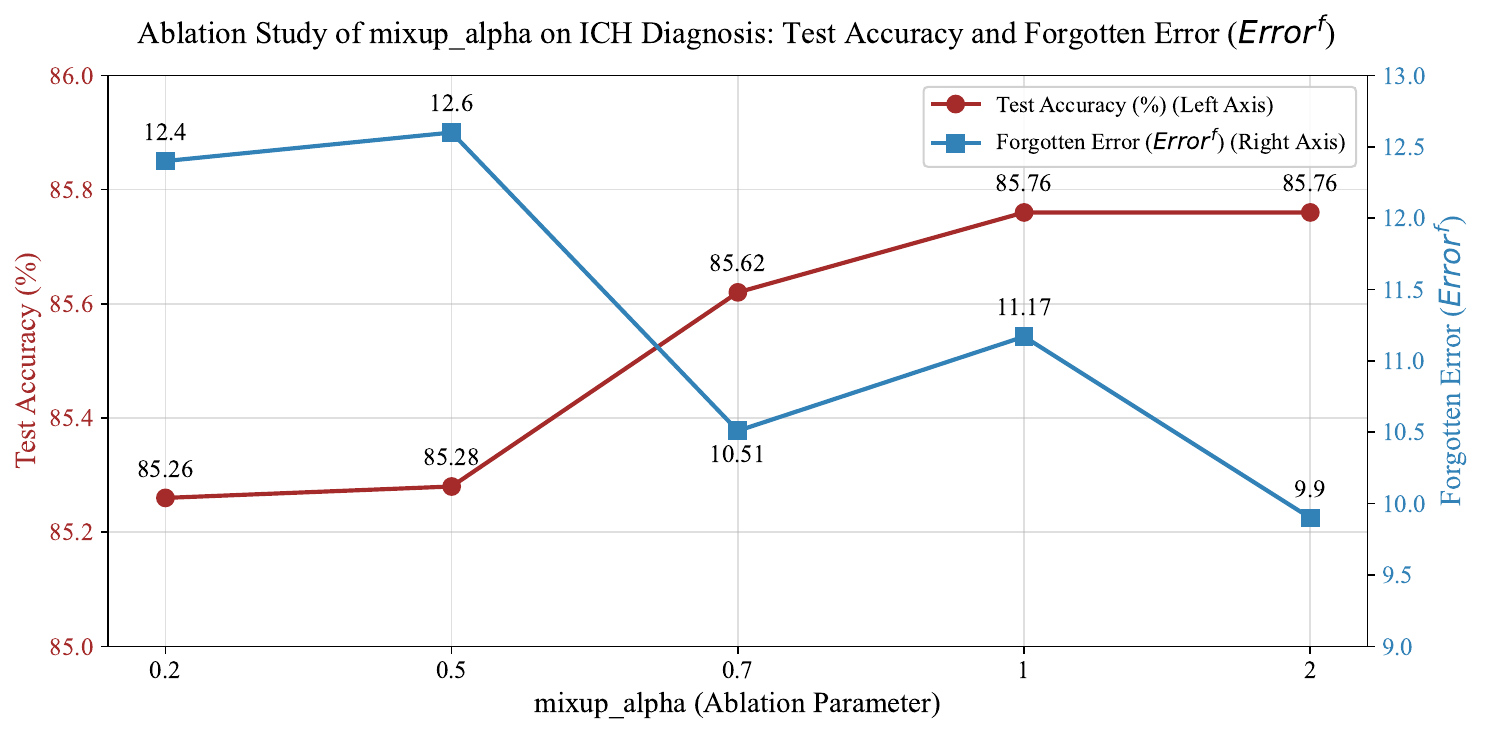} 
        \caption{Task 1 (ICH)} % 这里是图 (a) 的子标题
        \label{fig:ich_result}
    \end{subfigure}
    \hfill 
    
    \begin{subfigure}{\textwidth}
        \centering
        \includegraphics[width=\textwidth]{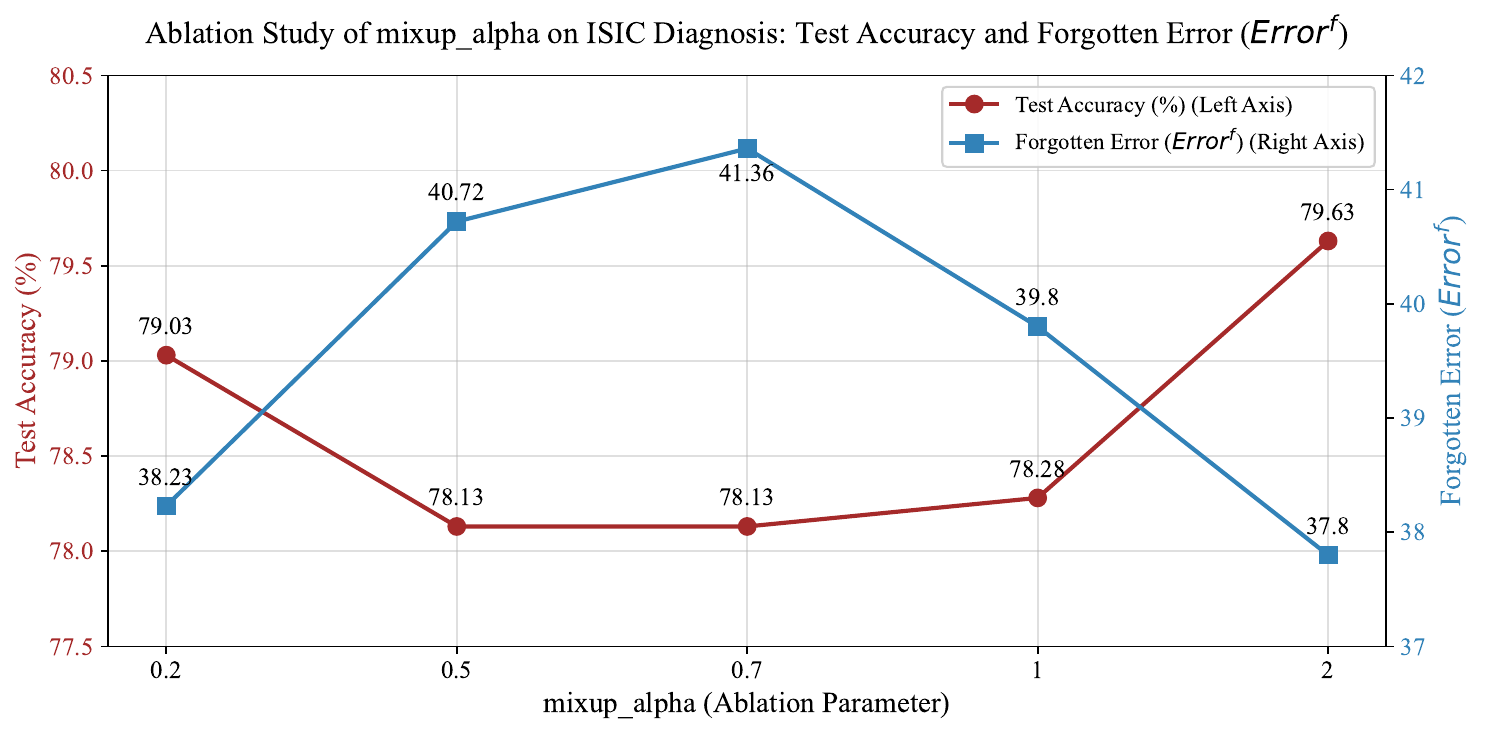}
        \caption{Task 2 (ISIC)}
        \label{fig:isic_result}
    \end{subfigure}
    
    % --- 整组图片的总标题 ---
    \caption{Ablation study of Image Feature Fusion Mechanism across mixup\_alpha, showing forgotten $Errorf$ and Accuracy for the two tasks.}
    \label{fig:ablation_study}
\end{figure}
\begin{table}[H]
\centering
\caption{The impact of applied probability on the results.}
\label{tab:3}
\begin{tabular}{c|cccc|c|c}
\toprule
\multirow{2}{*}{Benchmark and $p_{\text{mixup}}$} & \multicolumn{4}{c|}{Fidelity} & \multicolumn{1}{c|}{Efficacy} & \multicolumn{1}{c}{Efficiency} \\
\cmidrule{2-7}
& Accuracy & F1 & $Error^t$ & $Error^r$ & $Error^f$ & Runtime (s) \\
\midrule
ICH\cite{flanders2020rsna}(0.25) & 84.86 & 84.78 & 15.13 & 8.81 & 12.80 & 72 \\
ICH\cite{flanders2020rsna}(0.50) & 85.26 & 85.21 & 14.7 & 8.52 & 12.40 & 245 \\
ICH\cite{flanders2020rsna}(0.75) & 85.04 & 84.96 & 14.95 & 8.87 & 12.65 & 85 \\
ICH\cite{flanders2020rsna}(1) & 85.02 & 84.96 & 14.98 & 8.40 & 12.60 & 195 \\
\midrule
ISIC2018\cite{tschandl2018ham10000}(0.25) & 79.08 & 51.31 & 20.91 & 11.43 & 38.01 & 99 \\
ISIC2018\cite{tschandl2018ham10000}(0.50) & 79.03 & 51.30 & 20.96 & 11.62 & 38.23 & 120 \\
ISIC2018\cite{tschandl2018ham10000}(0.75) & 79.53 & 52.47 & 20.46 & 11.28 & 37.30 & 171 \\
ISIC2018\cite{tschandl2018ham10000}(1) & 78.58 & 51.31 & 21.41 & 11.34 & 37.37 & 72 \\
\bottomrule
\end{tabular}
\end{table}
\subsection{Ablation}

Since our approach inherently integrates the Image Feature Fusion Mechanism into the baseline Federated Client Unlearning (FCU)\cite{deng2024enable} framework, the comparative experiments presented in Table~1 and Table~2 have already demonstrated the effectiveness of our added mechanism (i.e., yielding superior unlearning capabilities compared to the vanilla FCU\cite{deng2024enable}). However, to more comprehensively understand how it specifically functions, we conduct a two-pronged ablation study dissecting the mechanism from both the mixing intensity (governed by $\alpha$) and the execution probability ($p_{\text{mixup}}$) perspectives.

\textbf{1. Controllability of the Mechanism and Dynamic Reshaping of Unlearning Boundaries.}
As shown in Figure~2, the dynamic trends of Test Accuracy and forgotten error ($\text{Error}^f$) across different $\alpha$ values confirm that the Image Feature Fusion Mechanism is not a passive black box, but rather deeply intervenes in the reconstruction process of the model's feature space. Serving as a ``control knob,'' $\alpha$ determines the intensity of the linear fusion between the forgotten samples and their surrounding retained samples, thereby fundamentally altering the model's unlearning boundary.

\textbf{2. Cross-Task Dynamic Trade-off and Adaptive Capability.}
The two tasks exhibit distinct yet regular trade-off dynamics, reflecting the mechanism's adaptability to varying data distributions:
\begin{itemize}
    \item \textbf{ICH\cite{flanders2020rsna} diagnostic task.} We see a clear trade-off: when you crank up the mixing intensity, fidelity and unlearning depth start moving in opposite directions. At $\alpha = 0.2$, the model hits an $\text{Error}^f = 12.4$, which is only $+0.12\%$ away from the retrained gold standard ($12.28$)---pretty much on target---while keeping good fidelity (Accuracy $85.26\%$). Bump $\alpha$ to $0.5$, and accuracy edges up by a tiny $0.02\%$, but you also get a slight drift toward over-forgetting ($\text{Error}^f = 12.6$). Once $\alpha \ge 0.7$, something we call \textbf{over-fusion} kicks in---the retain part's anchoring effect starts to dominate the mixed samples, and $\text{Error}^f$ drops sharply to $10.51\%$ or even lower. That means serious under-forgetting. So here's the takeaway: in binary medical diagnosis, the decision boundary is clean, and too much fusion just blurs it.

    \item \textbf{ISIC2018\cite{tschandl2018ham10000} diagnostic task.} This one shows an even sharper conflict between the two goals. $\text{Error}^f$ goes up then down (inverse U-shape), while accuracy goes down then up (U-shape). At $\alpha = 0.7$, the fusion mechanism aggressively destroys the target representations, pushing $\text{Error}^f$ to a peak of $41.36\%$. But that aggressive move also messes with global representations, causing accuracy to bottom out at $78.13\%$. On the flip side, at $\alpha = 2$ the model gives its best accuracy ($79.63\%$) with an $\text{Error}^f$ of $37.8$---that's only $+2.43\%$ above the retrained gold standard ($35.37\%$), which actually beats the default $\alpha = 0.2$ on this task in terms of alignment. This reveals an interesting \textbf{cross-task divergence}: ISIC2018\cite{tschandl2018ham10000} has a highly entangled multi-class feature manifold, so it seems to benefit from stronger feature fusion to untangle local client knowledge. Meanwhile, ICH\cite{flanders2020rsna} has a cleaner binary decision boundary and just gets hurt by over-fusion.
\end{itemize}

\textbf{3. Efficiency-Efficacy Trade-off via Execution Probability ($p_{\text{mixup}}$).}
$\alpha$ controls how strongly features get fused, while $p_{\text{mixup}}$ decides how often mixed samples show up in each training batch. Together, they give us two independent---and practically very important---knobs for tuning the mechanism. Table~3 breaks it down: as you vary $p_{\text{mixup}}$, you clearly see a \textbf{three-way trade-off} between fidelity, efficacy, and efficiency.
\begin{itemize}
    \item \textbf{On the ICH\cite{flanders2020rsna} benchmark,} raising $p_{\text{mixup}}$ from $0.25$ to $1.0$ gives diminishing returns. Unlearning depth barely improves ($\text{Error}^f$ goes from $12.80\%$ to $12.60\%$), and fidelity only edges up a bit (Accuracy from $84.86\%$ to $85.02\%$). But runtime jumps dramatically---from $72$ seconds all the way to $195$ seconds. What this suggests is: for datasets with relatively compact feature manifolds, a moderate $p_{\text{mixup}}$ (say, $0.75$) is enough to get most of the fusion mechanism's benefits, without paying the full computational price of universal mixing.
    
    \item \textbf{The ISIC2018\cite{tschandl2018ham10000} task shows a different kind of response behavior.} At $p_{\text{mixup}} = 0.75$, we get the best Pareto frontier---highest fidelity (Accuracy $79.53\%$, F1 $52.47\%$) and also the strongest unlearning efficacy ($\text{Error}^f = 37.30\%$). Interestingly, pushing $p_{\text{mixup}}$ all the way to $1.0$ doesn't help. It actually hurts fidelity (Accuracy drops to $78.58\%$), and $\text{Error}^f$ ($37.37\%$) barely moves beyond the $0.75$ setting, yet runtime crashes down to $72$ seconds. What's going on? Likely too much mixing noise in this higher-variance dermatoscopic dataset causes premature convergence. This kind of task-dependent behavior tells us one thing: adaptive configuration really matters. Just cranking up mixing to the max isn't always the best move.
\end{itemize}

Taking all the ablation results together, the core value of the Image Feature Fusion Mechanism seems reasonably clear. It not only tends to give the model stronger unlearning ability than the baseline FCU\cite{deng2024enable} in our experiments, but also offers a flexible dual‑parameter control space for federated unlearning—covering both the $\alpha$ and the execution probability $p_{\text{mixup}}$By tuning these two coupled hyperparameters appropriately, one may adjust the trade‑off among unlearning depth (Efficacy), global fidelity (Fidelity), and wall‑clock efficiency (Efficiency), depending on task complexity and hardware constraints of clinical edge devices, potentially reaching a favorable performance point. We choose $\alpha=0.2$ and $p_{\text{mixup}}=0.5$ as the default configuration, mainly to prioritize stability across different tasks. Why? Because while $\alpha=2$ appears to give better alignment on ISIC2018\cite{tschandl2018ham10000}, it leads to noticeable under‑forgetting on ICH\cite{flanders2020rsna} (error on forget set drops to 9.9, which could be problematic). The more conservative $\alpha=0.2$ achieves near‑perfect gold‑standard matching on ICH\cite{flanders2020rsna} and still provides acceptable alignment on ISIC2018\cite{tschandl2018ham10000}, thus avoiding task‑specific failure modes.
\section{Discussion}
\subsection{Interpretation of Findings and Comparison with Previous Studies}
The goal of machine unlearning is not to blindly maximize \(Error_f\), but to approximate the behavioral distribution of a model retrained from scratch on the retained data. Existing methods tend to fail at opposite ends of this objective. MoDe\cite{zhao2023federated} and UPGA\cite{halimi2022federated} 
achieve aggressive erasure by repelling isolated forget samples, but this forcefully tears apart adjacent retained representations, producing catastrophic fidelity collapse (deviations of \(+15.5\%\) and \(+12.9\%\) on ISIC2018\cite{tschandl2018ham10000}).Conversely,FFMU\cite{che2023fast} and Finetune lean toward under-forgetting, leaving residual target memorization that undermines regulatory compliance.IFF-FCU addresses this by reframing unlearning as a manifold boundary regularization problem. Rather than targeting discrete forget points, the Mixup-based Image Feature Fusion mechanism synthesizes a continuous transition zone between the forget and retain distributions, encouraging the model to generalize its forgetting behavior across the entire boundary region. This is theoretically grounded in recent work demonstrating that interpolated samples can simulate the entangled manifold regions that drive catastrophic forgetting.Empirically, IFF-FCU achieves an \(Error_f\) deviation of only \(+0.12\%\) from the retrained gold standard on ICH—the closest alignment among all evaluated methods—and \(+2.86\%\) on ISIC2018\cite{tschandl2018ham10000}, 
while maintaining competitive fidelity throughout. Runtime is reduced to 245s and 120s respectively, roughly an order of magnitude below full retraining, and a fraction of the cost of UPGA\cite{halimi2022federated} and FUKD\cite{wu2022federated}.
\subsection{Limitations of the Current Work}
First, mixed sample construction introduces modest computational 
overhead relative to FCU\cite{deng2024enable}, which may warrant further optimization in resource-constrained clinical edge deployments.

Second, the framework is sensitive to the mixing intensity parameter \(\alpha\), whose optimal value diverges across tasks. This implies dataset-specific tuning is required when deploying on new medical datasets, motivating future work on adaptive or meta-learned hyperparameter strategies.

Third, the cross-client retain sample access used in our experiments warrants clarification. This design was adopted solely to construct a controlled mixed distribution for rigorous empirical validation, and is not a fundamental requirement of the framework. In practice, the target client can construct mixed samples entirely from its own local data, retrieving retain samples that are semantically adjacent to the forget samples via embedding-space similarity search. Since 
the boundary regularization effect of Mixup arises from the 
interpolation between forget-proximal and retain-proximal 
representations rather than from their cross-client origin, such 
locally sourced samples serve as functionally equivalent anchors, preserving full compliance with federated privacy constraints.

Finally, evaluation is confined to image classification. Extending IFF-FCU to dense prediction tasks such as lesion segmentation and object detection—where spatial feature entanglement is considerably more complex—remains an important open direction.
\section{Conclusion}
In this paper, we propose IFF-FCU, a novel federated unlearning framework. The core innovation of our method lies in the sample-level Mixup augmentation strategy, which performs linear fusion between each forgotten sample and its surrounding retained samples, theoretically enabling the model to expand the boundary of forgetting,thereby achieving better unlearning efficacy.
\subsection{Future Work}
Several directions remain worth exploring further:

\begin{itemize}
    \item \textbf{Adaptive Sample Pairing:} Currently, the retained samples are selected through random sampling or index matching. A more sophisticated strategy could leverage similarity metrics in the input space or embedding space to pair each forgotten sample with the most ``influential'' retained sample, potentially enhancing the boundary learning between forgetting and retention.
    
    \item \textbf{Dynamic Mixup Scheduling:} Currently, the mixing coefficient $\lambda$ is sampled from a fixed Beta distribution. We could design a dynamic scheduling mechanism that gradually decreases the mixing strength during the unlearning process, allowing the model to transition smoothly from ``coarse-grained boundary exploration'' to ``fine-grained forgetting refinement''.
\end{itemize}
\textbf{Author Contributions:} Conceptualization, H.S. and W.J.; 
methodology, H.S. and G.S.; software, H.S. and Y.P.; 
validation, H.S., G.S. and T.L.; formal analysis, H.S. and G.S.; 
investigation, Y.P. and T.L.; writing---original draft preparation, H.S.; 
writing---review and editing, G.S., Y.P., T.L. and W.J.; 
visualization, Y.P. and T.L.; supervision, W.J.

\noindent \textbf{Funding:}This research received no external funding.

\noindent \textbf{Institutional Review Board Statement:}Not applicable

\noindent \textbf{Informed Consent Statement:}Not applicable

\noindent \textbf{Data Availability Statement:}The ISIC2018 data sets are openly available at https://www.kaggle.com/datasets/shonenkov/isic2018.And
the RSNA-ICH dataset are openly available at https://www.kaggle.com/c/rsna-intracranial-hemorrhage-detection.

\noindent \textbf{Conflicts of Interest:}The authors declare no conflict of interest. 

\vspace{1em} % 留出一点垂直间距
\noindent \textbf{Abbreviations} \\
The following abbreviations are used in this manuscript: \\[0.5em]

\noindent
\begin{tabular}{@{}ll}
CCPA    & California Consumer Privacy Act \\
FCU     & Federated Client Unlearning \\
FedAvg  & Federated Averaging \\
FL      & Federated Learning \\
FU      & Federated Unlearning \\
GDPR    & General Data Protection Regulation \\
IFF-FCU & Image Feature Fusion-based Federated Client Unlearning \\
\end{tabular}
\bibliography{references}

@inproceedings{deng2024enable,
  title={Enable the Right to be Forgotten with Federated Client Unlearning in Medical Imaging},
  author={Deng, Zhipeng and Luo, Luyang and Chen, Hao},
  booktitle={International Conference on Medical Image Computing and Computer-Assisted Intervention (MICCAI)},
  year={2024}
}

@article{flanders2020rsna,
  title={Construction of a machine learning dataset through collaboration: the rsna 2019 brain ct hemorrhage challenge},
  author={Flanders, Adam E and Prevedello, Luciano M and Shih, George and Halabi, Safwan S and Kalpathy-Cramer, Jayashree and Ball, R and Mongan, John T and Stein, A and Kitamura, Felipe C and Lungren, Matthew P and others},
  journal={Radiology: Artificial Intelligence},
  volume={2},
  number={3},
  pages={e190211},
  year={2020},
  publisher={Radiological Society of North America}
}

@article{tschandl2018ham10000,
  title={The HAM10000 dataset, a large collection of multi-source dermatoscopic images of common pigmented skin lesions},
  author={Tschandl, Philipp and Rosendahl, Cliff and Kittler, Harald},
  journal={Scientific data},
  volume={5},
  number={1},
  pages={1--9},
  year={2018},
  publisher={Nature Publishing Group}
}

@inproceedings{mcmahan2017communication,
  title={Communication-efficient learning of deep networks from decentralized data},
  author={McMahan, Brendan and Moore, Eider and Ramage, Daniel and Hampson, Seth and y Arcas, Blaise Ag{\"u}era},
  booktitle={Artificial intelligence and statistics},
  pages={1273--1282},
  year={2017},
  organization={PMLR}
}

@inproceedings{che2023fast,
  title={Fast Federated Machine Unlearning with Nonlinear Functional Theory},
  author={Che, Tianshi and Zhou, Yang and Zhang, Zijie and Lyu, Lingjuan and Liu, Ji and Yan, Da and Dou, Dejing and Huan, Jun},
    
booktitle={Fortieth International Conference on Machine Learning
},
    year={2023}
}

@article{halimi2022federated,
  title={Federated unlearning: How to efficiently erase a client in fl?},
  author={Halimi, Anisa and Kadhe, Swanand and Rawat, Ambrish and Baracaldo, Nathalie},
  journal={arXiv preprint arXiv:2207.05521},
  year={2022}
}

@article{zhao2023federated,
  title={Federated unlearning with momentum degradation},
  author={Zhao, Yian and Wang, Pengfei and Qi, Heng and Huang, Jianguo and Wei, Zongzheng and Zhang, Qiang},
  journal={IEEE Internet of Things Journal},
  year={2023},
  publisher={IEEE}
}

@inproceedings{liu2021federaser,
  title={Federaser: Enabling efficient client-level data removal from federated learning models},
  author={Liu, Gaoyang and Ma, Xiaoqiang and Yang, Yang and Wang, Chen and Liu, Jiangchuan},
  booktitle={2021 IEEE/ACM 29th International Symposium on Quality of Service (IWQOS)},
  pages={1--10},
  year={2021},
  organization={IEEE}
}

@article{wu2022federated,
  title={Federated unlearning with knowledge distillation},
  author={Wu, Chen and Zhu, Sencun and Mitra, Prasenjit},
  journal={arXiv preprint arXiv:2201.09441},
  year={2022}
}

@article{voigt2017eu,
  title={The eu general data protection regulation (gdpr). A Practical Guide, 1st Ed., Cham},
  author={Voigt, P. and Von dem Bussche, A.},
  journal={Springer International Publishing},
  volume={10},
  number={3152676},
  pages={10--5555},
  year={2017}
}

@article{harding2019understanding,
  title={Understanding the scope and impact of the california consumer privacy act of 2018},
  author={Harding, E.L. and Vanto, J.J. and Clark, R. and Hannah Ji, L. and Ainsworth, S.C.},
  journal={Journal of Data Protection \& Privacy},
  volume={2},
  number={3},
  pages={234--253},
  year={2019}
}

@inproceedings{peng2025adversarial,
  title     = {Adversarial Mixup Unlearning},
  author    = {Peng, Zhuoyi and Tang, Yixuan and Yang, Yi},
  booktitle = {The 2025 International Conference on Learning Representations (ICLR)},
  year      = {2025},
  address   = {Singapore},
  month     = {April},
  eprint    = {2502.10288},
  archivePrefix = {arXiv},
  primaryClass = {cs.LG},
  url       = {https://openreview.net/forum?id=GcbhbZsgiu}
}

@misc{wang2026jellyfishzeroshotfederatedunlearning,
      title={Jellyfish: Zero-Shot Federated Unlearning Scheme with Knowledge Disentanglement}, 
      author={Houzhe Wang and Xiaojie Zhu and Chi Chen},
      year={2026},
      eprint={2604.04030},
      archivePrefix={arXiv},
      primaryClass={cs.CR},
      url={https://arxiv.org/abs/2604.04030}, 
}

@inproceedings{
    zhong2025unlearning,
    title={Unlearning through Knowledge Overwriting: Reversible Federated Unlearning via Selective Sparse Adapter},
    author={Zhengyi Zhong and Weidong Bao and Ji Wang and Shuai Zhang and Jingxuan Zhou and Lingjuan Lyu and Wei Yang Bryan Lim},
    booktitle={Proceedings of the IEEE/CVF conference on Computer Vision and Pattern Recognition (CVPR)},
    year={2025},
    url={https://openreview.net/forum?id=3EUHkKkzzj}
}

@inproceedings{sheng2024robust,
  title     = {Robust Federated Unlearning},
  author    = {Sheng, Xinyi and Bao, Wei and Ge, Liming},
  booktitle = {Proceedings of the 33rd ACM International Conference on Information and Knowledge Management (CIKM)},
  pages     = {2034--2044},
  year      = {2024},
  publisher = {ACM},
  doi       = {10.1145/3627673.3679817},
  url       = {https://doi.org/10.1145/3627673.3679817}
}

@misc{maher2026fedzmugan,
  title        = {FedzMuGAN: Zero-Shot Class-Level Machine Unlearning in Federated Settings Using Generative Adversarial Networks},
  author       = {Maher, Mohamed and Mahfouz, Yara and others},
  year         = {2026},
  publisher    = {SSRN},
  doi          = {10.2139/ssrn.6389519},
  url          = {https://papers.ssrn.com/sol3/papers.cfm?abstract_id=6389519}
}
\end{document}